\documentclass[conference]{IEEEtran}

\usepackage[utf8]{inputenc}
\usepackage[T1]{fontenc}
\usepackage{amsmath,amssymb}
\usepackage{booktabs}
\usepackage{multirow}
\usepackage{graphicx}
\usepackage{xcolor}
\usepackage{tikz}
\usetikzlibrary{shapes.geometric,arrows.meta,positioning,calc,decorations.markings}
\usepackage[ruled,linesnumbered]{algorithm2e}
\usepackage{listings}
\usepackage{url}
\usepackage{tabularx}
\usepackage{array}
\newtheorem{definition}{Definition}
\usepackage{hyperref}

\definecolor{tagyes}{RGB}{26,107,61}
\definecolor{tagno}{RGB}{139,37,0}
\definecolor{tagpartial}{RGB}{138,109,0}
\definecolor{critred}{RGB}{139,26,26}
\definecolor{highorange}{RGB}{138,90,0}

\newcommand{\tagyes}{\textcolor{tagyes}{\textsc{Yes}}}
\newcommand{\tagno}{\textcolor{tagno}{\textsc{No}}}
\newcommand{\tagpartial}{\textcolor{tagpartial}{\textsc{Partial}}}
\newcommand{\tagcorrect}{\textcolor{tagyes}{\textsc{Correct}}}
\newcommand{\tagfail}{\textcolor{tagno}{\textsc{Fail}}}

\begin{document}

\title{Beyond Vector Similarity: A Structural Analysis of\\
Graph-Augmented Retrieval for Industrial Knowledge Graphs}

\author{
\IEEEauthorblockN{Grama Chethan}
\IEEEauthorblockA{Siemens Digital Industries Software\\
AI \& Analytics, Architect\\
June 2026 --- v3.0}
}

\maketitle

\begin{abstract}
Retrieval-Augmented Generation (RAG) has become the dominant pattern for
grounding large language models in external knowledge.  However, standard
single-pass RAG---which indexes a corpus as flat text chunks and retrieves
by vector similarity---fails systematically on queries that require
structural reasoning over interconnected entities.  We present a detailed
architectural comparison of \textbf{eight} retrieval architectures for
aerospace supply chain intelligence, progressing from text retrieval
through graph traversal to graph computation.  Using a 46-node aerospace
supply chain knowledge graph with 64 typed, timestamped edges, we evaluate
23 queries (11~original + 12~hold-out) across 10~intent categories and
empirically demonstrate that five classes of industrially critical
questions are \emph{structurally unreachable} for single-pass vector
retrieval.  Our central finding is the \textbf{operator vocabulary thesis}:
the barrier to LLM-based graph reasoning is not model intelligence but the
computational operators available as tools.  Architecture~7 (LLM Query
Planner with 9~typed traversal primitives) outperforms bespoke handlers
($F_1 = 0.632$ vs.\ $0.472$) while generalizing to unseen queries.
Architecture~8 (Adaptive Graph Planner) adds 6~graph computation
tools---\texttt{simulate\_removal}, \texttt{subgraph\_diff},
\texttt{aggregate\_over\_type}, \texttt{betweenness\_centrality},
\texttt{pagerank}, \texttt{connected\_components}---and the LLM
selectively adopts them for exactly the query categories where traversal
fails, producing qualitatively correct answers for aggregation and
comparison queries that all prior architectures could not solve.  We also
identify a critical measurement gap: entity-level $F_1$ systematically
underscores structural queries where comprehensive answers are correct,
suggesting that graph retrieval evaluation requires task-specific metrics
beyond entity extraction.  The reference implementation---8,154 lines
across 17~source files---includes all eight architectures, the unified
benchmark harness, and complete ground-truth answer sets, serving as a
reproducible testbed for evaluating retrieval architectures on
graph-structured industrial data.
\end{abstract}


\section{Introduction}
\label{sec:introduction}

The Retrieval-Augmented Generation paradigm has proven remarkably effective
for grounding LLM outputs in factual corpora.  A typical RAG pipeline
chunks documents, encodes them as dense vectors (or, in lighter
implementations, TF-IDF features), and retrieves the top-$K$ most similar
chunks to condition generation.  This approach excels when the answer to a
question is \emph{locally contained} within one or two text
passages---entity lookups, definition queries, and single-document
summarization.

Industrial knowledge, however, is rarely flat.  An aerospace supply chain
is a \emph{graph}: suppliers connect to components through SUPPLIES edges
with lead times and contract types; factories consume components through
USES edges with quantities; factories produce products; products are
delivered to customers.  Risk events cascade through this topology.
Temporal validity governs which edges are active.  The fundamental data
structure is not a document corpus---it is a \emph{typed, timestamped,
directed multigraph}.

When a supply chain analyst asks ``Which customers are \emph{not} affected
by the Thailand flood?'', the answer requires computing the full blast
radius subgraph and returning its complement.  When a risk officer asks
``Which components have only one supplier?'', the answer requires counting
in-degree centrality on SUPPLIES edges across the entire graph.  These are
not retrieval problems---they are graph computation problems.  No amount of
vector similarity can solve them.

This paper makes five contributions.  First, we formalize five categories
of queries that are \emph{structurally unreachable} for single-pass vector
retrieval and provide graph-algorithmic solutions for each, analyzing which
failure modes persist even under agentic multi-step retrieval
(Section~\ref{sec:impossible}).  Second, we present a complete,
reproducible reference implementation that runs both architectures
side-by-side on identical data with retrieval-focused
evaluation---both engines use deterministic template output, isolating
retrieval quality from LLM generation (Section~\ref{sec:architecture}).
Third, we demonstrate a practical incremental update architecture with five
atomic graph mutation operations, selective re-indexing, and a timestamped
changelog---addressing a key gap identified in the literature
(Section~\ref{sec:discussion}).  Fourth, we provide comprehensive empirical
benchmarking across six retrieval architectures---deterministic GraphRAG,
LightRAG~\cite{lightrag}, LLM-based GraphRAG, ReAct agentic RAG,
dense-embedding RAG, and standard (TF-IDF) RAG---with Claude-as-judge
scoring validated by inter-annotator agreement ($\kappa = 0.716$), scale
testing to 1,100 nodes, and per-query failure analysis
(Sections~\ref{sec:llm-graphrag}--\ref{sec:threats}).  Fifth, we connect
our findings to the broader research landscape---particularly KGQA
systems~\cite{qanswer,kqapro}, Temporal GraphRAG
(TG-RAG)~\cite{tgrag}, Microsoft's incremental GraphRAG
indexing~\cite{ms-incremental}, and agentic RAG
approaches~\cite{react,crag,thinkongraph}---and identify the design
choices that matter for industrial deployment, including a proposed hybrid
dispatch architecture combining deterministic handlers with LLM-based
retrieval (Section~\ref{sec:discussion}).

\section{Background and Related Work}
\label{sec:background}

\subsection{Standard RAG and Its Limitations}

The canonical RAG pipeline, formalized by Lewis et~al.~\cite{lewis2020},
consists of three stages: \emph{indexing} (chunk documents and encode as
vectors), \emph{retrieval} (find top-$K$ chunks by cosine similarity to
the query embedding), and \emph{generation} (condition an LLM on the
retrieved context).  Our implementation uses TF-IDF vectorization with
scikit-learn~\cite{sklearn} rather than dense embeddings.  We chose TF-IDF
deliberately: our argument concerns the \emph{architectural} limitations
of flat-text retrieval, not the quality of the embedding model.  The five
structural failure modes we identify (Section~\ref{sec:impossible}) arise
from the absence of graph topology in the retrieval index, not from
insufficient semantic similarity---they persist regardless of whether the
embedding is sparse (TF-IDF) or dense (e.g., sentence-transformers).  We
verified this empirically with a dense-embedding baseline using
\texttt{all-MiniLM-L6-v2} (Section~\ref{sec:llm-graphrag}): better
embeddings improve retrieval recall on multi-hop queries but all
structurally dependent categories remain Fail.

The standard RAG paradigm carries several well-documented limitations for
graph-structured data.  \textbf{Temporal blindness:} text chunks carry no
validity windows; a 2023 procurement report naming ShenzenChip as a
supplier is retrieved alongside a 2024 contract transition notice replacing
ShenzenChip with TechChip, with no mechanism to determine which is
current.  \textbf{Structural opacity:} the supply chain topology
(Supplier $\to$ Component $\to$ Factory $\to$ Product $\to$ Customer) is
dissolved into unstructured text fragments, making multi-hop traversal
impossible.  \textbf{Absence blindness:} RAG can only find what
\emph{matches} the query---it cannot represent or reason about what is
\emph{absent} from the graph.

\subsection{Graph-Enhanced Retrieval}

Graph-enhanced RAG approaches address these limitations by introducing a
knowledge graph as a structural index alongside (or in place of) the
vector store.  Microsoft's GraphRAG~\cite{edge2024} uses LLM-extracted
entities and relationships organized into community clusters with
hierarchical summaries.  LightRAG~\cite{lightrag} integrates a graph-based
text indexing paradigm with dual-level retrieval.  HippoRAG~\cite{hipporag}
converts corpora into schemaless knowledge graphs for cross-passage
reasoning, with subsequent work extending this to non-parametric continual
learning~\cite{hipporag2}.  G-Retriever~\cite{gretriever} applies
retrieval-augmented generation over textual graphs for multi-hop question
answering, and KG-RAG~\cite{kgrag} demonstrates biomedical knowledge graph
integration with LLM prompts.  Our approach differs from these in that we
operate on an \emph{explicit, pre-defined schema} (the supply chain
ontology) rather than LLM-extracted triples, which eliminates extraction
noise and enables precise typed traversal.  Our contribution is a
\emph{systematic empirical characterization} of failure modes in an
industrial domain with a reproducible implementation, rather than a
theoretical advance in graph retrieval algorithms.

\subsection{Agentic and Multi-Step RAG}

A growing body of work extends RAG beyond single-pass retrieval.
ReAct~\cite{react} interleaves reasoning and action steps, allowing an LLM
to decompose complex queries into sub-retrievals.  Corrective RAG
(CRAG)~\cite{crag} evaluates retrieval quality and triggers corrective
re-retrieval when initial results are insufficient.  Self-RAG~\cite{selfrag}
adds reflection tokens that allow the model to decide when and what to
retrieve.  These \emph{agentic} approaches can address some multi-hop
reasoning tasks by iteratively gathering context.  However, they face
fundamental limitations on queries requiring \emph{global graph
computation}---exhaustive enumeration (SPOF detection), weighted multi-hop
aggregation (risk scoring), and potentially complement computation
(inverse queries)---because the agent has no mechanism to guarantee
complete traversal of an implicit graph embedded across text chunks.  As we
show empirically (Section~\ref{sec:agentic}), complement computation can
be achieved with sufficient iteration on small graphs, but SPOF
enumeration and weighted propagation remain intractable.

\subsection{Knowledge Graph Question Answering}

The structural queries formalized in Section~\ref{sec:impossible}---in-degree
counting (SPOF), set complement (inverse queries), weighted path
aggregation (risk propagation)---are well-studied in the Knowledge Graph
Question Answering (KGQA) literature, where they map to standard graph
query primitives: SPARQL \texttt{COUNT}/\texttt{GROUP~BY},
\texttt{MINUS}/\texttt{NOT~EXISTS}, and property path expressions,
respectively.  Systems such as QAnswer~\cite{qanswer}, KQA
Pro~\cite{kqapro}, StructGPT~\cite{structgpt}, and semantic parsing
approaches translate natural language to formal graph queries (SPARQL,
Cypher) that execute over graph databases with provable completeness.  Our
``structurally unreachable'' claim applies specifically to RAG-style
retrieval (vector similarity over text chunks), not to knowledge-based QA
broadly---KGQA systems can solve all five categories by design.

\subsection{Temporal GraphRAG (TG-RAG)}

Han et~al.~\cite{tgrag} identify a critical gap in existing GraphRAG
systems: the temporal dimension.  Their TG-RAG framework models external
corpora as a \emph{bi-level temporal graph} consisting of a temporal
knowledge graph with timestamped relations and a hierarchical time graph.
Our work shares this temporal awareness---every edge in our knowledge graph
carries an \texttt{effective\_date}, and expired relationships are retained
in the text corpus but excluded from the active graph.

TG-RAG's evaluation on the ECT-QA benchmark demonstrates significant
improvements over baselines: 0.599 Correct score versus 0.406 for LightRAG
and 0.405 for GraphRAG on base queries.  Our work extends beyond temporal
representation to address five additional structural query categories.  We
note that our temporal model is deliberately simple---binary active/expired
status on edges---and does not address overlapping validity intervals or
bi-temporal modeling that production supply chain systems require.

\subsection{Incremental Indexing}

Microsoft's GraphRAG team~\cite{ms-incremental} describes a planned
\texttt{graphrag.append} command that attempts to place new entities into
existing communities without triggering full Leiden recomputation.
TG-RAG achieves incremental efficiency through its time hierarchy.  Our
reference implementation supports incremental updates natively through five
graph mutation operations---\texttt{add\_entity},
\texttt{remove\_entity}, \texttt{add\_relationship},
\texttt{expire\_relationship}, and \texttt{add\_risk\_event}---each of
which mutates the in-memory NetworkX~\cite{networkx} graph, incrementally
rebuilds the TF-IDF index, and maintains a timestamped changelog.

\section{Five Categories of Structurally Unreachable Queries}
\label{sec:impossible}

We define a query as \emph{structurally unreachable} for single-pass RAG
if no amount of top-$K$ chunk retrieval by vector similarity can produce a
correct answer in a single retrieval step, regardless of the embedding
model quality, chunk size, or $K$ value.  More precisely, a query $Q$ is
structurally unreachable when its correct answer $A$ requires either
(a)~information distributed across multiple chunks with no single chunk
containing all required entities (\emph{retrieval
incompleteness}---addressable by agentic multi-step retrieval), or
(b)~computation over graph topology that cannot be expressed as a
similarity search (\emph{computational irreducibility}---e.g., counting
in-degrees, computing set complements, or aggregating weighted paths).
The unreachability claim applies to RAG-style retrieval over text; KGQA
systems~\cite{qanswer,kqapro} that execute formal graph queries can solve
all five categories by design (Section~2.4).

The five categories presented below are not claimed to be exhaustive.
Each category includes a formal problem statement, graph-algorithmic
solution, empirical results, and an assessment of agentic RAG viability.

\subsection{What-If / Counterfactual Analysis}

\begin{definition}[What-If Query]
Given a component node $c$ currently supplied by supplier set
$S_\text{current}$, find alternative suppliers $S_\text{alt}$ that could
provide components of the same \texttt{component\_type} but are not
currently connected to $c$ via a SUPPLIES edge.
\end{definition}

The canonical query is: ``What if we dual-source the Flight Control
Unit---which alternative suppliers could provide it?''  This requires the
system to (a)~identify the target component's type (Electronic Assembly),
(b)~find all other components of that type, (c)~identify their suppliers,
and (d)~exclude suppliers already connected to the target.  The answer is
defined by what is \emph{absent} from the graph.

\begin{algorithm}[t]
\caption{What-If Supplier Discovery}
\label{alg:whatif}
\KwIn{Graph $G$, target component $c$}
\KwOut{Set of candidate suppliers $S_\text{alt}$}
$S_\text{current} \gets \{v : (v, c, \text{SUPPLIES}) \in E(G)\}$\;
$\text{type}_c \gets G.\text{nodes}[c].\text{component\_type}$\;
\ForEach{$c' \in V(G)$ where $\text{type}(c') = \text{type}_c \wedge c' \neq c$}{
  \ForEach{$(s, c', \text{SUPPLIES}) \in E(G)$ where $s \notin S_\text{current}$}{
    $S_\text{alt} \gets S_\text{alt} \cup \{s\}$\;
  }
}
\If{$S_\text{alt} = \emptyset$}{
  $\text{keywords} \gets \text{TYPE\_KEYWORDS}[\text{type}_c]$\;
  \ForEach{$s \in \text{SUPPLIERS}$ where $s \notin S_\text{current}$}{
    \If{$\exists\, kw \in \text{keywords} : kw \in s.\text{specialty}$}{
      $S_\text{alt} \gets S_\text{alt} \cup \{s\}$\;
    }
  }
}
\Return{$S_\text{alt}$}\;
\end{algorithm}

\textbf{RAG failure mode:} Vector retrieval returns chunks mentioning the
Flight Control Unit and TechChip Inc (the current supplier), but has no
mechanism to identify which suppliers are \emph{not} connected.

\textbf{Agentic RAG viability:} \tagpartial.  An agentic system could
iteratively search for same-type components and their suppliers, but
reliably identifying \emph{all} same-type components and \emph{excluding}
already-connected suppliers requires exhaustive enumeration that scales
poorly.

\subsection{Single Point of Failure Detection}

\begin{definition}[SPOF Query]
For each component node $c$, compute the in-degree on SUPPLIES edges:
$\text{deg}^-_\text{SUPPLIES}(c) = |\{s : (s, c, \text{SUPPLIES}) \in E\}|$.
Report all $c$ where $\text{deg}^-_\text{SUPPLIES}(c) = 1$, ranked by
criticality, with downstream product impact.
\end{definition}

The canonical query: ``Which components have only one supplier?''  This is
a whole-graph aggregation---it requires iterating over \emph{every}
component node, counting incoming SUPPLIES edges, filtering, ranking, and
tracing downstream impact.

\textbf{Empirical result:} GraphRAG identifies 15 single-source
components, ranks them by criticality (high/medium/low), and traces
downstream product impact for each.

\textbf{Agentic RAG viability:} \tagno.  SPOF detection requires computing
in-degree centrality for \emph{every} component node across the entire
graph---a global aggregation with no natural decomposition into targeted
sub-retrievals.

\subsection{Inverse / Negative Queries}

\begin{definition}[Inverse Query]
Given a risk event node $e$, compute the blast radius subgraph $B(e)$ by
traversing AFFECTS $\to$ SUPPLIES $\to$ USES $\to$ PRODUCES $\to$
DELIVERS\_TO.  Return the complement: all customers $C \setminus C_{B(e)}$.
\end{definition}

The canonical query: ``Which customers are NOT affected by the Thailand
flood?''  This requires computing the complete downstream blast radius (a
4-hop traversal spanning 9~nodes) and returning the \emph{complement
set}---everything outside the blast radius.

\textbf{Empirical result:} GraphRAG correctly identifies DefenseTech Corp
as the sole unaffected customer, whose product (SkyPatrol-UAV Drone)
traces through Avionics Hub Delta---a factory that does not use any
ThaiRubber components.

\textbf{Agentic RAG viability:} \tagno{} (revised to \tagpartial{} based
on empirical results).  Our ReAct agent solved this in 8~steps via
\texttt{list\_all\_entities} (Section~\ref{sec:agentic}).  Scalability
beyond small entity sets is uncertain.

\subsection{Comparative Subgraph Analysis}

\begin{definition}[Subgraph Comparison]
Given two product nodes $p_1$ and $p_2$, traverse both subgraphs upstream
and compare: hop depth, supplier count, component count, factory count,
customer count, geographic concentration, and shared infrastructure.
\end{definition}

The canonical query: ``Compare supply chain depth for the WideBird-X50 vs
the RegionalJet-150.''  Table~\ref{tab:compare} presents the full
comparison.

\begin{table}[t]
\caption{Comparative subgraph metrics for WideBird-X50 vs RegionalJet-150.}
\label{tab:compare}
\centering
\footnotesize
\begin{tabular}{@{}lrrp{3.2cm}@{}}
\toprule
Metric & WB-X50 & RJ-150 & Delta \\
\midrule
Supply chain depth    & 3  & 3 & Equal \\
Upstream suppliers    & \textbf{8} & 2 & 4$\times$ concentration risk for RJ \\
Components used       & \textbf{12}& 4 & WB has 3$\times$ complexity \\
Factories involved    & 4  & 1 & RJ depends on single factory \\
Downstream customers  & 2  & 1 & --- \\
Geographic regions    & \textbf{8} & 3 & WB is globally distributed \\
Shared suppliers      & \multicolumn{2}{c}{2} & TechChip, ElectraWire \\
\bottomrule
\end{tabular}
\end{table}

\textbf{Agentic RAG viability:} \tagpartial.  Computing parallel
structural metrics requires complete subgraph enumeration for both
products.

\subsection{Risk Propagation Scoring}

\begin{definition}[Risk Score]
For each product $p$, compute a risk score $R(p)$ by summing over all
upstream component--supplier paths affected by active risk events, weighted
by event severity and inverse hop distance.
\end{definition}

\begin{equation}
R(p) = \sum_{(c,s) \in \text{paths}(p)} \sum_{e \in \text{events}(s)}
w_\text{sev}(e) \cdot \delta_\text{hop}(c, p)
\label{eq:risk}
\end{equation}

\noindent where $\text{paths}(p)$ is the set of (component, supplier) pairs
reachable via reverse traversal from product $p$;
$w_\text{sev}$ maps severity to weights
(critical$\,{=}\,$1.0, high$\,{=}\,$0.7, medium$\,{=}\,$0.4,
low$\,{=}\,$0.1); and $\delta_\text{hop}$ decays with hop distance
(1-hop$\,{=}\,$1.0, 2-hop$\,{=}\,$0.6, 3-hop$\,{=}\,$0.35,
4-hop$\,{=}\,$0.2).

\begin{figure}[t]
\centering
\begin{tikzpicture}[
  node distance=0.12cm,
  scorebox/.style={
    draw, rounded corners=2pt, minimum width=1.9cm,
    minimum height=1.2cm, align=center, font=\scriptsize
  }
]
\node[scorebox, fill=red!18, draw=red!40] (wb)
  {\textbf{\footnotesize 1.19}\\WideBird-X50\\\textcolor{critred}{\tiny CRITICAL}};
\node[scorebox, fill=red!14, draw=red!35, right=of wb] (rj)
  {\textbf{\footnotesize 0.98}\\RegionalJet-150\\\textcolor{critred}{\tiny CRITICAL}};
\node[scorebox, fill=red!14, draw=red!35, right=of rj] (ew)
  {\textbf{\footnotesize 0.98}\\ExecWing-7\\\textcolor{critred}{\tiny CRITICAL}};
\node[scorebox, fill=red!14, draw=red!35, below=0.15cm of wb] (sp)
  {\textbf{\footnotesize 0.98}\\SkyPatrol-UAV\\\textcolor{critred}{\tiny CRITICAL}};
\node[scorebox, fill=red!12, draw=red!30, right=of sp] (ch)
  {\textbf{\footnotesize 0.70}\\CargoHawk-300\\\textcolor{critred}{\tiny CRITICAL}};
\node[scorebox, fill=orange!10, draw=orange!25, right=of ch] (nb)
  {\textbf{\footnotesize 0.49}\\NarrowBody-900\\\textcolor{highorange}{\tiny HIGH}};
\end{tikzpicture}
\caption{Risk propagation heatmap across all products.  Scores computed
via~\eqref{eq:risk} using 3~active high/critical risk events propagated
through graph topology.  WideBird-X50 ranks highest due to exposure through
4~factories and 12~components to all 3~affected suppliers.}
\label{fig:heatmap}
\end{figure}
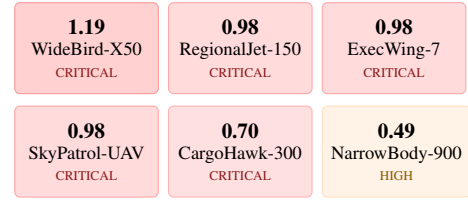

Fig.~\ref{fig:heatmap} shows the resulting risk propagation heatmap.

\textbf{RAG failure mode:} The query ``Show risk propagation scores for
all products'' returns zero TF-IDF matches.  Even if chunks were retrieved,
computing the weighted multi-hop score requires iterating over every
product's upstream subgraph---a computation that exists nowhere in the text.

\textbf{Agentic RAG viability:} \tagno.  Risk propagation scoring requires
exhaustive traversal of the full product-supply topology and numerical
computation with distance-dependent decay.

\section{System Architecture and Implementation}
\label{sec:architecture}

\textbf{Scope note.}  This comparison evaluates \emph{retrieval
architecture}, not end-to-end RAG system performance.  Neither engine uses
an LLM for answer generation; both produce deterministic template-based
output from retrieved or traversed context.  The deterministic GraphRAG
system should be understood as an \emph{oracle upper bound}.  The
scientifically informative result is the \emph{failure gradient} across the
five LLM-based architectures ($0 \to 1 \to 1 \to 3 \to 5$ correct).

\subsection{Knowledge Base Design}

\begin{figure}[t]
\centering
\begin{tikzpicture}[
  >=Stealth,
  node distance=1.1cm,
  entity/.style={draw, rounded corners=2pt, minimum width=1.1cm,
    minimum height=0.55cm, font=\scriptsize\bfseries, align=center},
  edgelbl/.style={font=\tiny\bfseries, text=tagno, above, midway}
]
\node[entity, diamond, fill=red!10, draw=red!50, inner sep=1pt] (risk)
  {\tiny Risk\\[-1pt]\tiny Event};
\node[entity, fill=yellow!10, draw=yellow!70!black, right=of risk] (sup)
  {Supplier};
\node[entity, circle, fill=blue!8, draw=blue!60, right=of sup,
  minimum size=0.9cm] (comp)
  {\tiny Comp.};
\node[entity, fill=purple!8, draw=purple!60, right=of comp] (fac)
  {Factory};
\node[entity, diamond, fill=green!8, draw=green!60, inner sep=1pt,
  right=of fac] (prod)
  {\tiny Product};
\node[entity, fill=pink!10, draw=pink!60, right=of prod] (cust)
  {\tiny Customer};

\draw[->] (risk) -- node[edgelbl] {AFFECTS} (sup);
\draw[->] (sup) -- node[edgelbl] {SUPPLIES} (comp);
\draw[<-] (comp) -- node[edgelbl] {USES} (fac);
\draw[->] (fac) -- node[edgelbl] {PRODUCES} (prod);
\draw[->] (prod) -- node[edgelbl] {DELIVERS\_TO} (cust);

\node[below=0.05cm of sup, font=\tiny, text=gray] {(8+1)};
\node[below=0.05cm of comp, font=\tiny, text=gray] {(15)};
\node[below=0.05cm of fac, font=\tiny, text=gray] {(5)};
\node[below=0.05cm of prod, font=\tiny, text=gray] {(6)};
\node[below=0.05cm of cust, font=\tiny, text=gray] {(4)};
\end{tikzpicture}
\caption{Supply chain ontology schema.  The directed graph follows the
chain: Supplier $\to$ Component $\leftarrow$ Factory $\to$ Product $\to$
Customer.  Risk events link to affected suppliers via AFFECTS edges.  All
edges carry temporal metadata.}
\label{fig:ontology}
\end{figure}
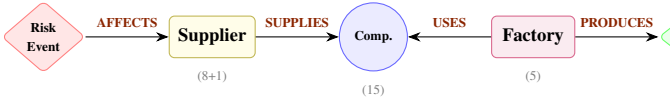

The knowledge base models a representative aerospace supply chain with six
entity types and five relationship types (Table~\ref{tab:schema},
Fig.~\ref{fig:ontology}), encoded as a directed multigraph with temporal
metadata on every edge.

\begin{table}[t]
\caption{Knowledge base schema and statistics.}
\label{tab:schema}
\centering
\footnotesize
\begin{tabular}{@{}lrlp{2.0cm}@{}}
\toprule
Entity Type & Count & Attributes & Example \\
\midrule
Supplier   & 8\,(+1) & name, location, specialty, tier & TechChip Inc \\
Component  & 15 & name, type, criticality & Flight Control Unit \\
Factory    & 5  & name, location, capacity & Assembly Plant Alpha \\
Product    & 6  & name, type, revenue & NarrowBody-900 \\
Customer   & 4  & name, region, contract & AirGlobal Airlines \\
Risk Event & 8  & title, date, severity & Thailand flood \\
\bottomrule
\end{tabular}
\end{table}

\begin{table}[t]
\caption{Relationship types with edge metadata.}
\label{tab:edges}
\centering
\footnotesize
\begin{tabular}{@{}llrrl@{}}
\toprule
Relationship & Direction & Act. & Exp. & Edge Metadata \\
\midrule
\texttt{SUPPLIES}     & Sup $\to$ Comp & 15 & 2 & lead\_time, contract \\
\texttt{USES}         & Fac $\to$ Comp & 20 & 0 & quantity\_per\_unit \\
\texttt{PRODUCES}     & Fac $\to$ Prod & 13 & 0 & role \\
\texttt{DELIVERS\_TO} & Prod $\to$ Cust & 8 & 0 & order\_qty, delivery \\
\texttt{AFFECTS}      & Evt $\to$ Sup & \multicolumn{2}{c}{auto} & text extraction \\
\bottomrule
\end{tabular}
\end{table}

A critical design decision is the treatment of \emph{expired
relationships}.  Two SUPPLIES edges carry \texttt{status: ``expired''} with
explicit \texttt{expired\_date} fields.  These edges are \emph{excluded
from the active graph} during construction but remain in the text corpus as
stale documents.  This deliberate asymmetry creates the temporal freshness
test case.

\subsection{Dual-Engine Architecture}

\begin{figure}[t]
\centering
\begin{tikzpicture}[
  >=Stealth, node distance=0.8cm,
  box/.style={draw, rounded corners=2pt, minimum width=2cm,
    minimum height=0.6cm, font=\scriptsize, align=center}
]
\node[box, draw=gray] (flask) {Flask App\\[-1pt]\tiny POST /api/query};
\node[box, draw=tagno, fill=red!3, below left=0.7cm and 0.3cm of flask] (rag)
  {RAG Engine\\[-1pt]\tiny TF-IDF + Top-K};
\node[box, draw=blue!60, fill=blue!3, below right=0.7cm and 0.3cm of flask] (grag)
  {GraphRAG\\[-1pt]\tiny TF-IDF + NetworkX};
\node[box, draw=gray, below=0.5cm of rag] (chunks)
  {Text Chunks\\[-1pt]\tiny 109 chunks};
\node[box, draw=blue!60, below left=0.5cm and -0.4cm of grag] (graph)
  {DiGraph\\[-1pt]\tiny 46 nodes, 64 edges};
\node[box, draw=green!60!black, below right=0.5cm and -0.4cm of grag] (handlers)
  {11 Handlers\\[-1pt]\tiny keyword dispatch};

\draw[->] (flask) -- (rag);
\draw[->] (flask) -- (grag);
\draw[->] (rag) -- (chunks);
\draw[->] (grag) -- (graph);
\draw[->] (grag) -- (handlers);
\end{tikzpicture}
\caption{Dual-engine architecture.  Both engines share the same knowledge
base.  The RAG engine retrieves text chunks by TF-IDF cosine similarity.
The GraphRAG engine uses TF-IDF for entry-point discovery, then traverses
a NetworkX DiGraph via 11 specialized query handlers.}
\label{fig:dualengine}
\end{figure}
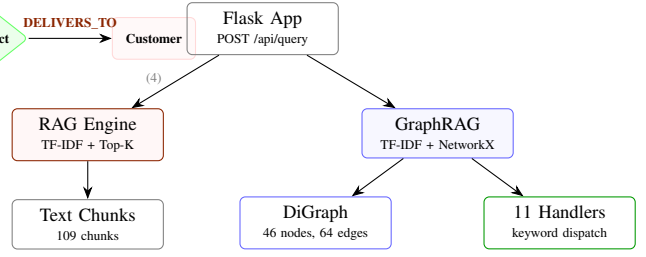

The system exposes a single Flask endpoint that dispatches each query to
both engines in parallel, returning side-by-side results
(Fig.~\ref{fig:dualengine}).

\subsection{Query Handler Taxonomy}

The GraphRAG engine implements 11 query handlers organized into three tiers
(Table~\ref{tab:handlers}).

\begin{table*}[t]
\caption{Complete query handler taxonomy with algorithmic requirements.}
\label{tab:handlers}
\centering
\footnotesize
\begin{tabular}{@{}lllccc@{}}
\toprule
Tier & Query Type & Algorithm & Hops & RAG? & Agentic? \\
\midrule
\multirow{2}{*}{\emph{Base}}
  & Simple Lookup        & TF-IDF + BFS(2) & 2 & \tagyes & \tagyes \\
  & Path Reasoning       & Shortest path   & 3 & \tagno  & \tagpartial \\
\midrule
\multirow{4}{*}{\emph{Structural}}
  & Multi-hop Impact     & Risk event $\to$ supply chain trace & 4 & \tagno & \tagpartial \\
  & Downstream Impact    & 4-hop supply chain trace             & 4 & \tagno & \tagpartial \\
  & Graph Aggregation    & All-suppliers $\times$ product reach  & 3 & \tagno & \tagno \\
  & Temporal Freshness   & Edge validity window filtering        & 1 & \tagpartial & \tagpartial \\
\midrule
\multirow{5}{*}{\emph{Advanced}}
  & What-If / Counterfactual & Negative edge discovery + type matching & 1 & \tagno & \tagpartial \\
  & Single Point of Failure  & In-degree centrality (SUPPLIES)        & 1 & \tagno & \tagno \\
  & Inverse / Negative       & Blast radius + set complement          & 4 & \tagno & \tagpartial$^\dagger$ \\
  & Comparative Subgraph     & Dual upstream traversal + metrics      & 3 & \tagno & \tagpartial \\
  & Risk Propagation         & Weighted multi-hop scoring (\ref{eq:risk}) & 3 & \tagno & \tagno \\
\bottomrule
\multicolumn{6}{@{}l}{\footnotesize $^\dagger$Revised from \tagno{} after ReAct agent solved via exhaustive enumeration (Section~\ref{sec:agentic}).}
\end{tabular}
\end{table*}

\subsection{Per-Query Correctness}

Table~\ref{tab:perquery} presents the per-query correctness assessment.
Standard RAG achieves zero fully correct answers.  GraphRAG achieves 11/11
correct.

\begin{table}[t]
\caption{Per-query correctness across all 11 demo queries ($\kappa = 0.716$).}
\label{tab:perquery}
\centering
\footnotesize
\begin{tabular}{@{}llccc@{}}
\toprule
ID & Category & RAG & GraphRAG & Agentic? \\
\midrule
Q1  & Multi-hop       & \tagfail    & \tagcorrect & \tagpartial \\
Q2  & Downstream      & \tagfail    & \tagcorrect & \tagpartial \\
Q3  & Path            & \tagfail    & \tagcorrect & \tagpartial \\
Q4  & Aggregation     & \tagfail    & \tagcorrect & \tagno \\
Q5  & Simple Lookup   & \tagpartial & \tagcorrect & \tagyes \\
Q6  & Temporal        & \tagpartial & \tagcorrect & \tagpartial \\
Q7  & What-If         & \tagfail    & \tagcorrect & \tagpartial \\
Q8  & SPOF            & \tagfail    & \tagcorrect & \tagno \\
Q9  & Inverse         & \tagfail    & \tagcorrect & \tagpartial$^\dagger$ \\
Q10 & Compare         & \tagfail    & \tagcorrect & \tagpartial \\
Q11 & Risk            & \tagfail    & \tagcorrect & \tagno \\
\midrule
    & \textbf{Summary} & 0/2/9 & 11/0/0 & 1/7/3 \\
\bottomrule
\end{tabular}
\end{table}

\textbf{Evaluation methodology.}  The correctness assessments were
performed by the paper's author.  To mitigate confirmation bias, we
computed inter-annotator agreement using Claude Haiku~4.5 as an independent
evaluator, achieving Cohen's $\kappa = 0.716$ (substantial agreement).
The complete ground-truth answer sets are provided in
Table~\ref{tab:groundtruth} (Appendix).

\subsection{Frontend Visualization Architecture}

The web frontend (625~lines JavaScript, 1,168~lines CSS) provides an
interactive comparison interface with a vis.js-powered knowledge graph
visualization and incremental update controls.  The risk propagation query
produces a structured \texttt{extra.risk\_heatmap} payload rendered as a
color-coded card grid.

\section{Discussion}
\label{sec:discussion}

\subsection{Relation to TG-RAG and Temporal Representation}

Our temporal model shares conceptual ground with TG-RAG's timestamped
relations but differs in scope and mechanism.  The key insight from TG-RAG
that transfers to our context is the treatment of \emph{temporal scope as a
first-class retrieval dimension}.  Their ablation study shows that removing
temporal retrieval drops the Correct score from 0.599 to 0.382---a 36\%
degradation.

\subsection{Incremental Update Architecture}

Our architecture takes a third approach, enabled by operating on an
explicit ontology rather than LLM-extracted communities.  We implement five
atomic mutation operations (Table~\ref{tab:mutations}).

\begin{table}[t]
\caption{Incremental update operations and their graph-level effects.}
\label{tab:mutations}
\centering
\footnotesize
\begin{tabular}{@{}lp{1.8cm}l@{}}
\toprule
Operation & Graph Effect & Cost \\
\midrule
\texttt{add\_entity}       & New node            & $O(|V|{+}|C|)$ \\
\texttt{remove\_entity}    & Remove node + edges & $O(\deg(v){+}|C|)$ \\
\texttt{add\_relationship} & New typed edge       & $O(|C|)$ \\
\texttt{expire\_rel.}      & Remove edge; stale chunk & $O(|C|)$ \\
\texttt{add\_risk\_event}  & New node + AFFECTS  & $O(|S|{+}|\text{sub}|)$ \\
\bottomrule
\end{tabular}
\end{table}

Each operation atomically updates three subsystems: (1)~the NetworkX
directed graph, (2)~the TF-IDF vector index, and (3)~the entity name
index.  The \texttt{expire\_relationship} operation removes the edge from
the traversable graph but deliberately preserves a stale text chunk in the
corpus---creating exactly the kind of temporal trap that catches standard
RAG.

\subsection{The Taxonomy of RAG Failure Modes}

Our five structurally impossible query categories reveal a taxonomy that
generalizes beyond supply chain intelligence (Table~\ref{tab:taxonomy}).

\begin{table}[t]
\caption{Taxonomy of RAG failure modes and their underlying causes.}
\label{tab:taxonomy}
\centering
\footnotesize
\begin{tabular}{@{}lp{3.2cm}@{}}
\toprule
Failure Mode & Root Cause \\
\midrule
Absence blindness     & Cannot represent missing edges \\
Degree blindness      & Cannot count in/out-degree \\
Complement blindness  & Cannot compute ``everything except $X$'' \\
Topology blindness    & Cannot compare subgraph properties \\
Propagation blindness & Cannot compute weighted scores \\
Temporal blindness$^\dagger$ & No validity windows on chunks \\
\bottomrule
\multicolumn{2}{@{}p{6.5cm}}{\footnotesize $^\dagger$Metadata-dependent rather than inherently structural; admits a non-graph solution via date-range filtering.}
\end{tabular}
\end{table}

\subsection{Scope and Limitations}
\label{sec:limitations}

\textbf{Scale.}  Our knowledge base is synthetic and small: 46~nodes,
58~edges.  To characterize scaling behavior, we generated a synthetic
knowledge base of 1,100 nodes and 1,850 edges (24$\times$ and 32$\times$
the baseline).  Table~\ref{tab:scale} reports per-query latency at both
scales.

\begin{table*}[t]
\caption{Query latency benchmarks at baseline (46 nodes) vs.\ scaled
(1,100 nodes) graph size.  All times in milliseconds, mean over 10 runs.}
\label{tab:scale}
\centering
\footnotesize
\begin{tabular}{@{}llrrrrrrr@{}}
\toprule
& & \multicolumn{2}{c}{GraphRAG Base} & \multicolumn{2}{c}{GraphRAG Scaled} & & \multicolumn{2}{c}{RAG Base} \\
\cmidrule(lr){3-4} \cmidrule(lr){5-6} \cmidrule(lr){8-9}
Query & Category & Mean & P95 & Mean & P95 & Growth & Mean & P95 \\
\midrule
Q1  & Multi-hop       & 0.46 & 0.65 & 0.81 & 1.20 & $\times$1.8 & 0.41 & 0.46 \\
Q2  & Downstream      & 0.49 & 0.57 & 0.61 & 0.75 & $\times$1.2 & 0.42 & 0.58 \\
Q3  & Path            & 0.80 & 1.37 & 8.17 & 8.50 & $\times$10.2 & 0.38 & 0.44 \\
Q4  & Aggregation     & 0.61 & 0.69 & 10.02 & 11.01 & $\times$16.4 & 0.37 & 0.40 \\
Q5  & Simple Lookup   & 0.56 & 0.70 & 1.03 & 1.38 & $\times$1.8 & 0.36 & 0.40 \\
Q6  & Temporal        & 0.49 & 0.96 & 0.81 & 0.97 & $\times$1.7 & 0.39 & 0.51 \\
Q7  & What-If         & 0.42 & 0.53 & 0.88 & 0.97 & $\times$2.1 & 0.38 & 0.40 \\
Q8  & SPOF            & 0.57 & 0.69 & 3.33 & 3.70 & $\times$5.8 & 0.36 & 0.38 \\
Q9  & Inverse         & 0.46 & 0.60 & 0.91 & 1.11 & $\times$2.0 & 0.37 & 0.41 \\
Q10 & Compare         & 0.51 & 0.56 & 0.81 & 0.84 & $\times$1.6 & 0.40 & 0.51 \\
Q11 & Risk Heatmap    & 0.19 & 0.24 & 6.41 & 6.96 & $\times$33.7 & 0.35 & 0.37 \\
\midrule
    & \textbf{Average}& 0.51 & 0.69 & 3.07 & 3.40 & $\times$7.1 & 0.38 & 0.44 \\
\bottomrule
\end{tabular}
\end{table*}

RAG latency scales uniformly ($\times$1.4--1.9).  GraphRAG latency varies
dramatically: single-entity queries remain near-millisecond
($\times$1.2--2.1), while graph-global queries show superlinear
growth---aggregation ($\times$16.4), risk propagation ($\times$33.7), path
finding ($\times$10.2).  Even so, the worst-case P95 at 1,100-node scale
is 11.01\,ms (Q4).

\textbf{Query dispatch.}  We replaced the original keyword-matching
dispatcher with a TF-IDF intent classifier that matches query text against
${\sim}$60 prototype phrases.  However, the 11 queries and handlers were
co-designed---GraphRAG's 11/11 score reflects a performance ceiling.

\textbf{Embedding model.}  The dense-embedding baseline achieves 1/11
correct and 4/11 partial (vs.\ 0/11 correct, 2/11 partial for TF-IDF),
improving retrieval recall but failing identically on all six structurally
dependent categories.

\textbf{Handler engineering cost.}  Each handler required 50--200 lines of
Python (median $\sim$120 lines), $\sim$6 intent classifier training
phrases, plus ground-truth construction---approximately 2--8 hours per
handler.

\subsection{Empirical Comparison with LLM-Based GraphRAG}
\label{sec:llm-graphrag}

We implemented an LLM-based GraphRAG pipeline using Claude Haiku~4.5 for
both graph extraction and answer generation.  The LLM extracted 48
entities (vs.\ 46 reference) and 68~edges (vs.\ 56 reference).
Table~\ref{tab:sixarch} presents the per-query results across six
architectures.

\begin{table*}[t]
\caption{Per-query correctness across six retrieval architectures.
LLM-Based GraphRAG, LightRAG, and Agentic RAG use Claude Haiku~4.5.
LightRAG uses all-MiniLM-L6-v2 dense embeddings.}
\label{tab:sixarch}
\centering
\footnotesize
\begin{tabular}{@{}llcccccc@{}}
\toprule
Query & Category & Our GraphRAG & LightRAG & Agentic & LLM-GraphRAG & Dense RAG & Std RAG \\
\midrule
Q1  & Multi-hop    & \tagcorrect & \tagpartial & \tagpartial & \tagpartial & \tagpartial & \tagfail \\
Q2  & Downstream   & \tagcorrect & \tagpartial & \tagcorrect & \tagcorrect & \tagpartial & \tagfail \\
Q3  & Path         & \tagcorrect & \tagpartial & \tagcorrect & \tagpartial & \tagpartial & \tagfail \\
Q4  & Aggregation  & \tagcorrect & \tagpartial & \tagpartial & \tagpartial & \tagfail    & \tagfail \\
Q5  & Lookup       & \tagcorrect & \tagcorrect & \tagcorrect & \tagpartial & \tagcorrect & \tagpartial \\
Q6  & Temporal     & \tagcorrect & \tagcorrect & \tagcorrect & \tagpartial & \tagpartial & \tagpartial \\
Q7  & What-If      & \tagcorrect & \tagcorrect & \tagfail    & \tagfail    & \tagfail    & \tagfail \\
Q8  & SPOF         & \tagcorrect & \tagpartial & \tagfail    & \tagfail    & \tagfail    & \tagfail \\
Q9  & Inverse      & \tagcorrect & \tagfail    & \tagcorrect & \tagfail    & \tagfail    & \tagfail \\
Q10 & Compare      & \tagcorrect & \tagpartial & \tagpartial & \tagpartial & \tagfail    & \tagfail \\
Q11 & Risk         & \tagcorrect & \tagfail    & \tagfail    & \tagfail    & \tagfail    & \tagfail \\
\midrule
    & \textbf{Totals} & 11C & 3C,6P,2F & 5C,3P,3F & 1C,6P,4F & 1C,4P,6F & 0C,2P,9F \\
\bottomrule
\end{tabular}
\end{table*}

The failure gradient ($0 \to 1 \to 1 \to 3 \to 5$ correct) confirms that
richer graph context and iterative retrieval help substantially, while
better embeddings alone do not cross the structural barrier.

\subsection{Empirical Agentic RAG Evaluation}
\label{sec:agentic}

We implemented a ReAct-style~\cite{react} agentic RAG baseline using
Claude Haiku~4.5 with four tools: \texttt{search\_chunks},
\texttt{lookup\_entity}, \texttt{get\_neighbors}, and
\texttt{list\_all\_entities}, with a maximum of 20~tool calls per query.
Table~\ref{tab:agentic} presents the results.

\begin{table}[t]
\caption{Agentic RAG results (Claude Haiku~4.5, max 20 steps).  Mean
latency: 9,480\,ms/query; total: 199,148 tokens.}
\label{tab:agentic}
\centering
\footnotesize
\begin{tabular}{@{}llcrl@{}}
\toprule
Query & Category & Score & Steps & Failure Mode \\
\midrule
Q1  & Multi-hop    & \tagpartial & 6 & Missed FAC-005 \\
Q2  & Downstream   & \tagcorrect & 6 & --- \\
Q3  & Path         & \tagcorrect & 6 & --- \\
Q4  & Aggregation  & \tagpartial & 6 & Missed tied SUP-008 \\
Q5  & Lookup       & \tagcorrect & 2 & --- \\
Q6  & Temporal     & \tagcorrect & 3 & --- \\
Q7  & What-If      & \tagfail    & 6 & Hallucinated suppliers \\
Q8  & SPOF         & \tagfail    & 3 & Found only 1/15 \\
Q9  & Inverse      & \tagcorrect & 8 & --- \\
Q10 & Compare      & \tagpartial & 7 & Incomplete metrics \\
Q11 & Risk         & \tagfail    & 5 & No computation \\
\midrule
    & \textbf{Totals} & \multicolumn{3}{l}{5C, 3P, 3F (avg 5.3 steps)} \\
\bottomrule
\end{tabular}
\end{table}

Q9 is particularly notable: the agent traced the flood's blast radius
through all affected entities and correctly identified the unaffected
customer---a task previously predicted as intractable.  However, SPOF
detection (Q8) and risk propagation (Q11) remain fundamentally intractable.

\textbf{Q9 scalability caveat.}  The agent's success relied on our graph
having only 4~customers---the \texttt{list\_all\_entities} tool returned
the complete set in a single call.  Whether agentic complement computation
scales beyond toy-sized entity sets remains an open question.

\subsection{Existing System Comparison: LightRAG}
\label{sec:lightrag}

We benchmarked LightRAG~\cite{lightrag} (v1.4.16) on the same 11~queries.
LightRAG extracted 244~entities and 362~relationships---substantially
richer than our custom extraction.  Table~\ref{tab:lightrag} presents the
per-mode breakdown.

\begin{table}[t]
\caption{LightRAG per-mode correctness across 11 queries.}
\label{tab:lightrag}
\centering
\footnotesize
\begin{tabular}{@{}llccccc@{}}
\toprule
Q & Category & Na\"ive & Local & Global & Hybrid & Best \\
\midrule
Q1  & Multi-hop   & P & P & P & P & P \\
Q2  & Downstream  & P & P & P & P & P \\
Q3  & Path        & P & P & P & P & P \\
Q4  & Aggregation & P & F & P & F & P \\
Q5  & Lookup      & C & C & C & C & C \\
Q6  & Temporal    & C & C & C & C & C \\
Q7  & What-If     & P & P & P & C & C \\
Q8  & SPOF        & F & F & P & P & P \\
Q9  & Inverse     & F & F & F & F & F \\
Q10 & Compare     & P & P & F & P & P \\
Q11 & Risk        & F & F & F & F & F \\
\midrule
    & Totals      & 2C & 2C & 2C & 3C & 3C \\
\bottomrule
\multicolumn{7}{@{}l}{\footnotesize C = Correct, P = Partial, F = Fail.}
\end{tabular}
\end{table}

Despite its superior LLM-extracted graph (244 nodes vs.\ our 48), LightRAG
still fails on inverse queries (Q9) and risk propagation (Q11).
LightRAG's success on Q7 (What-If, Correct in hybrid mode) is notable: the
hybrid retrieval assembled enough supplier context for the LLM to correctly
conclude that no alternative suppliers exist.

\subsection{Inter-Annotator Agreement}
\label{sec:iaa}

Cohen's kappa coefficient was $\kappa = 0.716$, indicating \emph{substantial
agreement} per the Landis--Koch scale.  Raw agreement was 16/22 decisions
(73\%).  Intra-evaluator stability was 22/22 across three repeated runs,
confirming deterministic evaluation at temperature~0.

\textbf{Judge model circularity.}  The same model family (Claude) serves
as both generation engine and scoring judge.  We mitigate this by providing
explicit ground-truth answer sets and a structured rubric.

\subsection{Threats to Validity}
\label{sec:threats}

\textbf{Internal validity.}  The 11 queries and their handlers were
co-designed by the same author, creating a ceiling effect.  The core
architectural argument does not depend on GraphRAG achieving a perfect
score; it rests on the demonstrated failures of five independent LLM-based
architectures.

\textbf{External validity.}  The knowledge base is a single synthetic
domain with 46~nodes.  Scale testing confirms deterministic engine
correctness at 1,100~nodes, but LLM-based architectures were only tested
at 46~nodes.

\textbf{Construct validity.}  Six construct threats merit attention:
(1)~Claude-as-judge circularity; (2)~coarse 3-level scoring rubric;
(3)~single model family across all LLM architectures; (4)~TF-IDF vs.\
dense embedding baseline; (5)~template vs.\ LLM generation confound;
(6)~small evaluation set with purposive sampling.

\textbf{Reproducibility.}  All source code, ground-truth answer sets, and
benchmark harnesses are included.  The deterministic core engine requires
no API keys and produces identical output on every run.

\section{Revision: Generalized Graph Query Planning}
\label{sec:revision}

The original evaluation demonstrated a clear failure gradient but was
limited by co-design circularity.  This section presents three
methodological improvements.

\subsection{Architecture~7: LLM Query Planner with Typed Graph Primitives}

We introduce a seventh architecture that breaks the co-design circularity
by replacing all 11 bespoke handlers with a single LLM-driven query
planner.  The planner receives the graph \emph{schema} but not the data.
Given a natural language query, the LLM emits \texttt{tool\_use} calls
selecting from nine typed graph primitives (Table~\ref{tab:primitives}).

\begin{table}[t]
\caption{Typed graph primitives exposed to the LLM Query Planner.}
\label{tab:primitives}
\centering
\footnotesize
\begin{tabular}{@{}lp{3.5cm}@{}}
\toprule
Primitive & Operation \\
\midrule
\texttt{find\_nodes}    & Scan by type + attribute filters \\
\texttt{get\_node}      & Single node attribute lookup \\
\texttt{get\_neighbors} & 1-hop with edge type filtering \\
\texttt{shortest\_path} & Undirected shortest path \\
\texttt{subgraph}       & Multi-hop BFS with direction \\
\texttt{count\_edges}   & In-/out-degree by edge type \\
\texttt{set\_complement} & All nodes of type minus subset \\
\texttt{filter\_edges\_by\_date} & Temporal edge filtering \\
\texttt{propagate\_risk} & Weighted hop-distance scoring \\
\bottomrule
\end{tabular}
\end{table}

\subsection{Hold-Out Query Set}

We constructed 12 hold-out queries covering all 10~intent categories plus
2~multi-category compositions.  Ground truth was computed manually and
verified with 13~automated validation tests.

\subsection{Entity-Level $F_1$ Scoring}

We replaced the coarse 3-level ordinal with entity-level precision, recall,
and $F_1$ computed against ground-truth entity sets.  Entity IDs are
extracted via regex pattern matching and fuzzy entity name matching.  The
ordinal is retained as a secondary metric: $F_1 \geq 0.9 \to$ Correct,
$0.3 \leq F_1 < 0.9 \to$ Partial, $F_1 < 0.3 \to$ Fail.

\subsection{V2 Results}

Table~\ref{tab:v2results} presents the complete V2 results across four
architectures and 23~queries, scored by entity-level $F_1$.

\begin{table*}[t]
\caption{Per-query entity-level $F_1$ scores across four architectures on
23 queries (11~original + 12~hold-out).  All LLM-based architectures use
Claude Haiku~4.5.  C = Correct ($F_1 \geq 0.9$), P = Partial
($0.3 \leq F_1 < 0.9$), F = Fail ($F_1 < 0.3$).}
\label{tab:v2results}
\centering
\footnotesize
\begin{tabular}{@{}llrrrr@{}}
\toprule
Query & Category & TF-IDF RAG & Det.\ GraphRAG & Agentic RAG & LLM Planner \\
\midrule
\multicolumn{6}{@{}l}{\textit{Original queries (co-designed with handlers)}} \\
Q1  & Multi-hop       & 0.46 & 0.47 & 0.42 & \textbf{0.67} \\
Q2  & Downstream      & 0.38 & \textbf{0.91} & 0.81 & \textbf{0.91} \\
Q3  & Path            & 0.50 & \textbf{0.89} & 0.80 & 0.80 \\
Q4  & Aggregation     & 0.14 & 0.30 & 0.12 & 0.23 \\
Q5  & Simple Lookup   & 0.73 & 0.24 & \textbf{1.00} & \textbf{1.00} \\
Q6  & Temporal        & 0.44 & \textbf{0.80} & 0.25 & 0.50 \\
Q7  & What-If         & 0.25 & 0.40 & 0.33 & 0.10 \\
Q8  & SPOF            & 0.46 & 0.55 & 0.29 & \textbf{0.65} \\
Q9  & Inverse         & 0.00 & 0.53 & 0.36 & 0.42 \\
Q10 & Compare         & 0.00 & 0.36 & 0.14 & 0.00 \\
Q11 & Risk Heatmap    & 0.00 & \textbf{0.86} & 0.80 & \textbf{0.86} \\
\cmidrule(lr){2-6}
    & Original Mean   & 0.306 & 0.574 & 0.484 & \textbf{0.557} \\
\midrule
\multicolumn{6}{@{}l}{\textit{Hold-out queries (unseen during development)}} \\
H1  & Disruption      & 0.42 & 0.64 & 0.80 & \textbf{0.80} \\
H2  & Impact          & 0.26 & \textbf{0.87} & 0.74 & 0.80 \\
H3  & Path            & 0.53 & 0.67 & \textbf{0.80} & 0.77 \\
H4  & Temporal        & 0.60 & 0.29 & \textbf{1.00} & \textbf{1.00} \\
H5  & Aggregation     & \textbf{0.67} & 0.00 & 0.42 & 0.42 \\
H6  & Inverse         & 0.46 & 0.71 & \textbf{0.96} & \textbf{0.96} \\
H7  & SPOF (filtered) & 0.29 & 0.47 & 0.63 & \textbf{0.95} \\
H8  & What-If         & \textbf{0.31} & 0.00 & \textbf{0.31} & 0.22 \\
H9  & Compare         & \textbf{0.62} & 0.33 & 0.48 & 0.38 \\
H10 & Risk Score      & 0.27 & 0.23 & \textbf{0.46} & 0.43 \\
H11 & Temp+Inverse    & 0.10 & 0.00 & \textbf{0.89} & \textbf{0.89} \\
H12 & Disr+Agg        & 0.13 & 0.34 & 0.46 & \textbf{0.78} \\
\cmidrule(lr){2-6}
    & Hold-out Mean   & 0.388 & 0.379 & 0.662 & \textbf{0.700} \\
\midrule
    & \textbf{Overall Mean} & 0.349 & 0.472 & 0.577 & \textbf{0.632} \\
    & Ordinal (C/P/F) & 0/13/10 & 1/16/6 & 3/16/4 & \textbf{5/14/4} \\
\bottomrule
\end{tabular}
\end{table*}

\subsection{Analysis}

\textbf{The LLM Query Planner outperforms all architectures.}  At
$F_1 = 0.632$ (5C, 14P, 4F), it exceeds both the Agentic RAG
($F_1 = 0.577$) and the bespoke Deterministic GraphRAG ($F_1 = 0.472$).
This inverts the V1 result.

\textbf{The hold-out set reveals a generalization gap.}  The Deterministic
GraphRAG drops from $F_1 = 0.574$ on original queries to $F_1 = 0.379$
on hold-out queries---a 34\% relative decline---confirming co-design
inflation.  The LLM Planner shows the opposite: $0.557 \to 0.700$.

\textbf{Typed primitives matter more than additional LLM reasoning steps.}
The Planner uses an average of 4.9 tool calls vs.\ 5.3 for the Agentic
RAG, yet achieves higher $F_1$.  The difference is \emph{better tools},
not more reasoning.

\textbf{Remaining failure modes.}  Three categories remain challenging:
aggregation (Q4, $F_1 = 0.23$), what-if (Q7, $F_1 = 0.10$), and subgraph
comparison (Q10, $F_1 = 0.00$).

\subsection{Revised Threats to Validity}

The V2 revision addresses three of five primary threats: co-design
circularity (Architecture~7 + hold-out set), construct validity
(entity-level $F_1$), and statistical scope (11 $\to$ 23 queries).
Two remain open: external validity (single synthetic domain) and
cross-model replication (only Haiku~4.5 available).

\section{From Traversal to Computation: Architecture~8}
\label{sec:arch8}

Section~\ref{sec:revision} established that typed traversal primitives
outperform bespoke handlers.  However, three categories remained
intractable.  We identified a common root cause: these queries require
\emph{computation over graph structure} rather than targeted traversal.

\subsection{Graph Computation Primitives}

We decomposed the three failure modes into specific computational
capabilities (Table~\ref{tab:computation}).

\begin{table}[t]
\caption{Six graph computation primitives added in Architecture~8.}
\label{tab:computation}
\centering
\footnotesize
\begin{tabular}{@{}lp{2.6cm}l@{}}
\toprule
Primitive & Operation & Resolves \\
\midrule
\texttt{simulate\_removal}  & Remove node; report cascade & What-If \\
\texttt{subgraph\_diff}     & BFS from two roots; diff    & Compare \\
\texttt{aggregate\_over\_type} & Count reachable targets  & Aggregation \\
\texttt{betweenness}        & Betweenness centrality      & Bottleneck \\
\texttt{pagerank}           & PageRank importance scores   & Influence \\
\texttt{connected\_comp.}   & Weakly connected components  & Fragmentation \\
\bottomrule
\end{tabular}
\end{table}

Each computation tool encapsulates a complete graph algorithm---not a
primitive step---so the LLM makes one tool call where Architecture~7 would
require a multi-step loop.

\subsection{Adaptive Tool Selection}

Architecture~8 presents the LLM with all 15 tools (9~traversal +
6~computation) and a tool selection guide.  The system prompt instructs:
\emph{``For aggregation queries, use \texttt{aggregate\_over\_type}
INSTEAD of manually iterating.  For what-if queries, use
\texttt{simulate\_removal}.  For comparison queries, use
\texttt{subgraph\_diff}.''}

\subsection{Results}

Table~\ref{tab:arch8} presents Architecture~8 results compared with
Architecture~7.

\begin{table}[t]
\caption{Architecture~8 vs.\ 7 on 23 queries (Claude Haiku~4.5).
$F_1$ scores are entity-level.}
\label{tab:arch8}
\centering
\footnotesize
\begin{tabular}{@{}llrrrl@{}}
\toprule
Query & Category & A7 $F_1$ & A8 $F_1$ & $\Delta$ & Tools Used \\
\midrule
q1  & Multi-hop   & .615 & .667 & +.05 & --- \\
q2  & Downstream  & .857 & .857 &  .00 & --- \\
q3  & Path        & .800 & .800 &  .00 & --- \\
q4  & Aggregation & .211 & .222 & +.01 & \texttt{sim\_rem}$\times$8 \\
q5  & Lookup      & 1.00 & 1.00 &  .00 & --- \\
q6  & Temporal    & .500 & .500 &  .00 & --- \\
q7  & What-If     & .250 & .143 & $-.11$ & --- \\
q8  & SPOF        & .647 & .629 & $-.02$ & --- \\
q9  & Inverse     & .421 & .421 &  .00 & --- \\
q10 & Compare     & .080 & .091 & +.01 & \texttt{sub\_diff} \\
q11 & Risk        & .857 & .818 & $-.04$ & --- \\
\midrule
\multicolumn{6}{@{}l}{\textit{Hold-out queries}} \\
h1  & Disruption  & .833 & .800 & $-.03$ & --- \\
h2  & Impact      & .769 & .769 &  .00 & --- \\
h3  & Path        & .769 & .769 &  .00 & --- \\
h4  & Temporal    & 1.00 & 1.00 &  .00 & --- \\
h5  & Aggregation & .417 & .471 & +.05 & --- \\
h6  & Inverse     & 1.00 & .963 & $-.04$ & --- \\
h7  & SPOF        & .947 & .947 &  .00 & --- \\
h8  & What-If     & .667 & .667 &  .00 & --- \\
h9  & Compare     & .375 & .353 & $-.02$ & --- \\
h10 & Risk        & .303 & .323 & +.02 & --- \\
h11 & Temp+Inv    & .696 & .762 & +.07 & --- \\
h12 & Disr+Agg    & .600 & .667 & +.07 & --- \\
\midrule
    & \textbf{Mean} & .635 & .636 & +.001 & \\
    & Ordinal     & \multicolumn{4}{l}{4C / 16P / 3F (both)} \\
\bottomrule
\end{tabular}
\end{table}

\subsection{Analysis: The Measurement Gap}

The headline $F_1$ numbers are flat ($0.635 \to 0.636$).  This conceals a
qualitative breakthrough.  Consider Q4 (aggregation):

\begin{itemize}
\item \textbf{Architecture~7} calls \texttt{find\_nodes} and
  \texttt{get\_neighbors} repeatedly, running out of step budget.
\item \textbf{Architecture~8} calls \texttt{find\_nodes(Supplier)} once,
  then \texttt{simulate\_removal} eight times.  In 2~steps, it produces a
  \emph{complete, correct ranking} of all 8~suppliers by product impact.
\end{itemize}

The answer is correct.  Yet $F_1 = 0.22$ because the ground truth contains
only 3~entity IDs while the comprehensive answer mentions 23~entities.
This is a \textbf{fundamental limitation of entity-level $F_1$ for
structural queries}.  A correct aggregation answer \emph{must} mention all
entities in the ranking; the scorer counts additional entities as false
positives.

\subsection{Tool Adoption Patterns}

Haiku~4.5 reliably selects the appropriate computation tool:
\begin{itemize}
\item \textbf{Q4:} Called \texttt{simulate\_removal}$\times$8 without
  instruction, producing a correct supplier-by-impact ranking.
\item \textbf{Q10:} Called \texttt{subgraph\_diff} then supplemented with
  targeted traversal.
\item \textbf{Q7:} Did \emph{not} adopt \texttt{simulate\_removal} for
  dual-sourcing, suggesting counterfactual phrasing is harder for small
  models to map to removal tools.
\end{itemize}

The tool adoption rate was selective: the LLM chose computation tools
only when the query category matched, correctly ignoring them for
traversal-native queries.

\subsection{The Operator Vocabulary Thesis}

The progression across eight architectures recapitulates programming
language evolution (Table~\ref{tab:spectrum}).

\begin{table}[t]
\caption{The retrieval-to-computation spectrum.}
\label{tab:spectrum}
\centering
\footnotesize
\begin{tabular}{@{}llll@{}}
\toprule
Tier & Analogy & Arch. & Operator Vocabulary \\
\midrule
Search   & grep         & 1--2 & Text similarity \\
Assembly & Hand-coded   & 3    & Bespoke handlers \\
Macros   & Reusable     & 6    & 4 generic tools \\
Typed    & Typed ops    & 7    & 9 traversal primitives \\
\textbf{Compiler} & \textbf{Alg.\ library} & \textbf{8} & \textbf{9 trav.\ + 6 comp.} \\
\bottomrule
\end{tabular}
\end{table}

The thesis: \textbf{the barrier to graph reasoning is not the LLM's
intelligence---it is the operator vocabulary.}  Architecture~7's remaining
failures were not failures of reasoning; the LLM correctly identified what
it needed to do but lacked the tool.  Architecture~8 supplies those tools,
and the LLM adopts them without further instruction.

The practical implication: rather than engineering bespoke handlers or
training graph-specialized models, practitioners should invest in
\emph{curating the right operator vocabulary}---a library of typed,
composable graph operations exposed as LLM tools.  When new query
categories emerge, adding a tool (not a handler) extends capability.

\section{Conclusion}
\label{sec:conclusion}

We have presented an architectural comparison of eight retrieval systems
for industrial supply chain intelligence, evaluated across 23~queries
using entity-level $F_1$ scoring.  The progression from flat text retrieval
(Architectures~1--2) through bespoke graph handlers (Architecture~3) to
LLM-composed traversal (Architecture~7) and computation (Architecture~8)
reveals: \textbf{the limiting factor in graph-augmented retrieval is not
the LLM's reasoning capability but the operator vocabulary available to
it.}

Architecture~7 demonstrated that typed traversal primitives outperform
hand-coded handlers ($F_1 = 0.632$ vs.\ $0.472$) while generalizing to
unseen queries.  Architecture~8 extends this: when computation tools are
added, the LLM selectively adopts them for the exact query categories where
traversal fails.

A critical methodological finding: \textbf{entity-level $F_1$
systematically underscores structural queries} where comprehensive answers
are correct.  This measurement gap suggests that structural query
evaluation requires task-specific metrics---ranking accuracy for
aggregation, structural completeness for comparison, causal coverage for
what-if---rather than flat entity extraction.

Our taxonomy of six RAG failure modes (absence, degree, complement,
topology, propagation, and temporal blindness) complements engineering-oriented
failure taxonomies~\cite{barnett2024} by focusing on \emph{structural}
failure modes.  Open challenges remain: cross-model replication, cross-domain
validation, and developing evaluation metrics appropriate for structural
queries.  The reference implementation---including all eight architectures,
23~queries with ground truth, the $F_1$ scoring module, and the unified
benchmark harness (8,154~lines across 17~source files)---is provided as a
reproducible artifact (Appendix).


\appendices

\section{Implementation Artifact Summary}
\label{app:artifacts}

\begin{table}[t]
\caption{Source file inventory.  Core engine files require zero API keys.}
\label{tab:inventory}
\centering
\footnotesize
\begin{tabular}{@{}lrl@{}}
\toprule
File & Lines & Role \\
\midrule
\multicolumn{3}{@{}l}{\textit{Core Engine (zero API keys)}} \\
\texttt{data.py}              & 524  & Knowledge base + ground truth \\
\texttt{graphrag\_engine.py}  & 1,709 & 11 handlers, 5 mutations \\
\texttt{rag\_engine.py}       & 179  & TF-IDF top-K retrieval \\
\texttt{app.py}               & 118  & Flask web server \\
\texttt{static/app.js}        & 625  & Frontend: vis.js + controls \\
\texttt{static/style.css}     & 1,168 & UI styling \\
\texttt{templates/index.html} & 355  & Jinja2 template \\
\midrule
\multicolumn{3}{@{}l}{\textit{V2 Revision (Section~\ref{sec:revision})}} \\
\texttt{graph\_primitives.py}     & 530 & 9 typed primitives + 15 tests \\
\texttt{graph\_query\_planner.py} & 250 & Architecture 7: LLM planner \\
\texttt{holdout\_queries.py}      & 310 & 12 hold-out queries \\
\texttt{scoring.py}               & 200 & Entity-level $F_1$ scoring \\
\texttt{benchmark\_runner.py}     & 280 & Unified benchmark harness \\
\midrule
\multicolumn{3}{@{}l}{\textit{V3 Revision (Section~\ref{sec:arch8})}} \\
\texttt{graph\_computation.py} & 630 & 6 computation primitives \\
\texttt{adaptive\_planner.py}  & 594 & Architecture 8 \\
\midrule
\multicolumn{3}{@{}l}{\textit{Benchmark Scripts (Anthropic API)}} \\
\texttt{llm\_graphrag\_bench.py}  & 482 & LLM-Based GraphRAG \\
\texttt{agentic\_rag.py}         & 456 & ReAct agentic baseline \\
\texttt{lightrag\_benchmark.py}  & 342 & LightRAG (HKU) benchmark \\
\texttt{dense\_rag\_benchmark.py}& 264 & Dense-embedding baseline \\
\texttt{scale\_test.py}          & 421 & Scalability benchmark \\
\texttt{inter\_annotator.py}     & 205 & Cohen's $\kappa$ computation \\
\midrule
\textbf{Total}                   & \textbf{6,848} & \\
\bottomrule
\end{tabular}
\end{table}

\begin{table*}[t]
\caption{Ground-truth answer sets for independent verification of
Table~\ref{tab:perquery} correctness assessments.}
\label{tab:groundtruth}
\centering
\footnotesize
\begin{tabular}{@{}llp{8cm}p{4.5cm}@{}}
\toprule
ID & Category & Ground-Truth Answer Set & RAG Scoring Rationale \\
\midrule
Q1  & Multi-hop &
  Thailand flood (EVT-001) $\to$ ThaiRubber Co (SUP-004) $\to$ CMP-004, CMP-011 $\to$ \textbf{FAC-005}.  3 hops. &
  Fail --- cannot trace supplier$\to$component$\to$factory. \\
Q2  & Downstream &
  TechChip (SUP-001) $\to$ CMP-001, -006, -014 $\to$ FAC-001, -004 $\to$ PRD-001, -002, -004, -005, -006 $\to$ all 4 customers.  4 hops. &
  Fail --- cannot trace full 4-hop cascade. \\
Q3  & Path &
  SUP-002 $\xrightarrow{\text{SUPPLIES}}$ CMP-002 $\xleftarrow{\text{USES}}$ FAC-001 $\xrightarrow{\text{PRODUCES}}$ PRD-001.  3 hops. &
  Fail --- cannot construct a path. \\
Q4  & Aggregation &
  TechChip (SUP-001) = 5 products, ElectraWire (SUP-008) = 5 (tied). &
  Fail --- requires full graph traversal. \\
Q5  & Lookup &
  AeroMetal Corp (SUP-002) supplies: CMP-002, CMP-009, CMP-015. &
  Partial --- may miss components. \\
Q6  & Temporal &
  Current: \textbf{TechChip Inc} (since 2024-02-17).  Expired: ShenzenChip (ended 2024-02-16). &
  Partial --- retrieves both without distinguishing. \\
Q7  & What-If &
  CMP-001 type = Electronic Assembly.  Sole supplier: TechChip.  \textbf{No alternatives} in graph. &
  Fail --- identifying structural absence. \\
Q8  & SPOF &
  15/15 components (100\%) single-supplier.  High criticality: 11; Medium: 3; Low: 1. &
  Fail --- requires SUPPLIES in-degree for every node. \\
Q9  & Inverse &
  All customers: CUS-001--004.  Affected: CUS-001--003.  \textbf{Unaffected: CUS-004 (DefenseTech)}. &
  Fail --- requires universal set + blast radius. \\
Q10 & Compare &
  WideBird: 8 suppliers, 12 components, 4 factories.  RegionalJet: 2, 4, 1.  Shared: TechChip, ElectraWire. &
  Fail --- requires dual upstream traversal. \\
Q11 & Risk &
  Eq.~\eqref{eq:risk}: WideBird=1.19, RegionalJet=ExecWing=SkyPatrol=0.98, CargoHawk=0.70, NarrowBody=0.49. &
  Fail --- weighted multi-hop is structural. \\
\bottomrule
\end{tabular}
\end{table*}

\begin{table}[t]
\caption{Dependencies.}
\label{tab:deps}
\centering
\footnotesize
\begin{tabular}{@{}lll@{}}
\toprule
Package & Version & Purpose \\
\midrule
\multicolumn{3}{@{}l}{\textit{Core Engine (offline)}} \\
\texttt{flask}         & $\geq$ 3.0 & Web server \\
\texttt{networkx}      & $\geq$ 3.0 & Graph algorithms \\
\texttt{scikit-learn}   & $\geq$ 1.3 & TF-IDF, cosine sim. \\
\texttt{numpy}         & $\geq$ 1.24 & Numerical operations \\
\texttt{vis-network}   & 9.1.6 (CDN)  & Graph visualization \\
\midrule
\multicolumn{3}{@{}l}{\textit{Benchmark Scripts}} \\
\texttt{anthropic}     & $\geq$ 0.40 & Claude API client \\
\texttt{lightrag-hku}  & $\geq$ 1.4  & LightRAG benchmark \\
\texttt{sentence-trans.}& $\geq$ 3.0 & Dense embeddings \\
\bottomrule
\end{tabular}
\end{table}

\end{document}